\PassOptionsToPackage{table}{xcolor}
\documentclass{article}
\usepackage{iclr2027_conference,times,amsmath,amssymb,booktabs,graphicx,microtype}
\usepackage[hidelinks]{hyperref}
\usepackage{url,float}
\usepackage{tabularx}
\definecolor{TableHead}{HTML}{DBE1EC}
\definecolor{TableTint}{HTML}{D0EEE4}
\hypersetup{pdftitle={Rules to Tools: Executable Checks for LLM Agents in Scientific Computing},pdfauthor={Jingjie Ning, Guojiang Zhao, Chen Xu, Shanshan Zhong, Xiaochuan Li, Ji Zeng, Guolin Ke},pdfkeywords={LLM agents, scientific computing, executable specifications, tool use, code repair}}
\title{Rules to Tools: Executable Checks\\for LLM Agents in Scientific Computing}
\iclrfinalcopy
\author{%
\makebox[\dimexpr\textwidth-2\tabcolsep\relax][c]{%
\begin{tabular}{c}
Jingjie Ning\textsuperscript{1}\thanks{Corresponding authors: Jingjie Ning
(\texttt{jening@cs.cmu.edu}) and Guolin Ke (\texttt{keg1@dp.tech}).}\quad
Guojiang Zhao\textsuperscript{2}\quad
Chen Xu\textsuperscript{1}\quad
Shanshan Zhong\textsuperscript{1}\\[2pt]
Xiaochuan Li\textsuperscript{1}\quad
Ji Zeng\textsuperscript{1}\quad
Guolin Ke\textsuperscript{2,*}\\[4pt]
{\normalfont\small
\textsuperscript{1}School of Computer Science, Carnegie Mellon University\quad
\textsuperscript{2}DP Technology}\\[2pt]
{\normalfont\footnotesize
\{jening, chenxu2, szhong2, xiaochu4, jizeng\}@cs.cmu.edu}\\
{\normalfont\footnotesize zhaogj@dp.tech\quad keg1@dp.tech}
\end{tabular}}%
}

\begin{document}
\maketitle
\lhead{Preprint}
\vspace{-.15in}
\begin{abstract}
Scientific coding agents receive equations, boundary conditions, and output
requirements in writing, then must assess the programs they revise. Rules to
Tools (R2T) supplies prepared executable checks of public scientific
requirements. Matched SciCode repair groups share written checks, starting
programs, model, and budgets; the tool group receives a callable implementation.
Across two task-ID cohorts, complete repair is 26/30 with text and 29/30 with
the prepared checks. Three task IDs favor tools, one favors text, and eleven tie.
The eight-ID cohort scores 13/16 versus 15/16, with a task-cluster bootstrap
95\% interval of $[-12.5,43.75]$ percentage points for the difference. The
larger shared-definition SciCode cohort ties at 13/24 per group. Five
development-exposed tasks with alternate starting programs score 3/10 versus
7/10. The tool group favors tasks 17, 77, and 11; initial checks flag task 17
and report no violation for tasks 77 and 11. Task 37 favors text and has no
initial reported violation.
A fresh source-through-Python arm
also reaches 15/16, matching the dedicated command's aggregate. In a matched
PDE comparison, detailed text scores 23/24 and checks score 24/24, with 31.2\%
lower reported model output for checks. Agent-side output savings vary by
cohort, while public CPU use rises in both task-ID cohorts. These results
measure task-dependent repair outcomes and agent-side costs with prepared checks.

\end{abstract}

\begin{figure}[!ht]
\centering
\includegraphics[width=\linewidth]{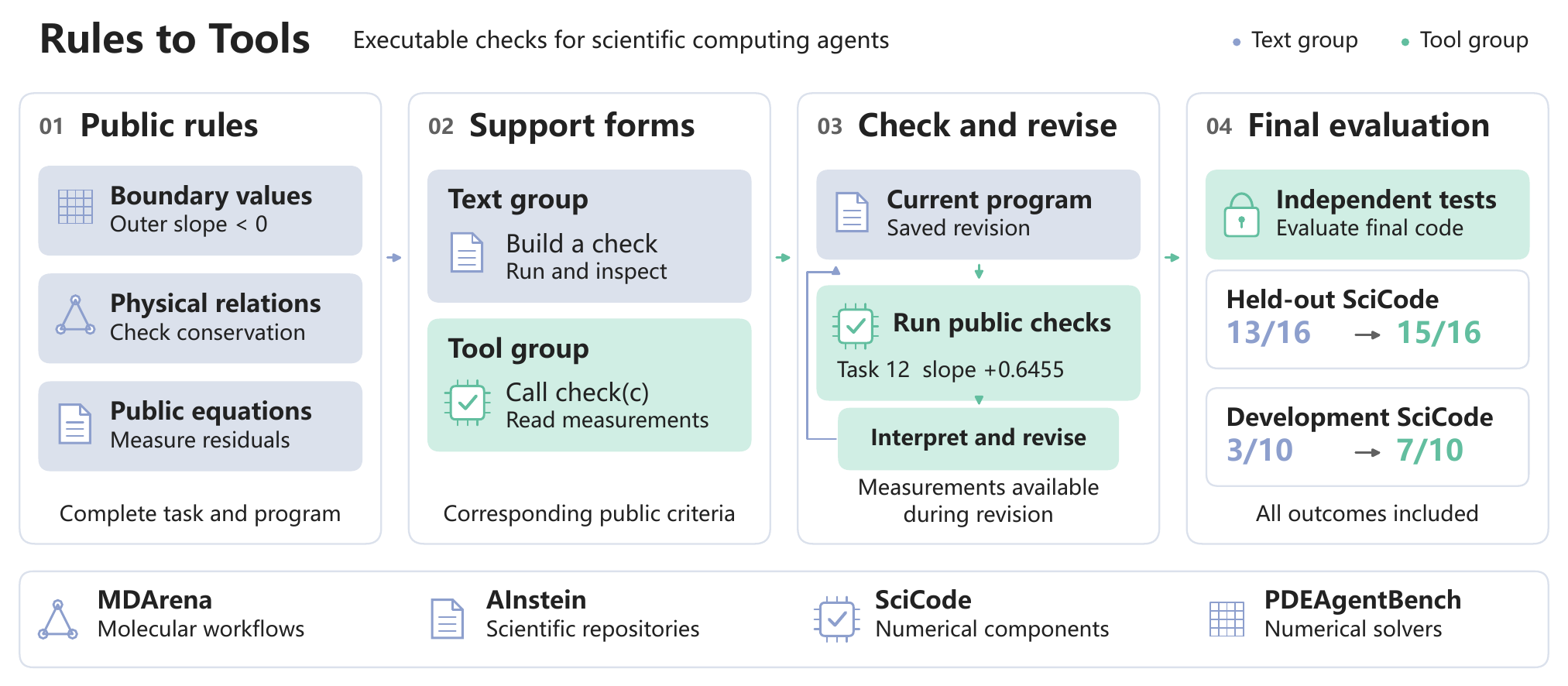}
\caption{Public scientific requirements support two verification workflows.
Corresponding written and executable specifications define the checks before
execution; measurements arise from running them on the current artifact.
The text group constructs checks through ordinary code execution. The tool
group can obtain structured measurements from a supplied implementation.
Both revise their artifacts and receive independent final evaluation.
Section~\ref{sec:design} details each group's information access and execution settings.}
\label{fig:live-checker-method}
\end{figure}

\section{Introduction}

Scientific computing uses programs to study physical systems and solve
mathematical problems. AI coding agents help by writing, running, and revising
these programs. Success requires both working code and results that satisfy the
scientific requirements. A simulation may finish running while producing values
that violate a physical law or disagree with a prescribed equation.
Scientific coding benchmarks test these demands through numerical calculations,
molecular simulations, and scientific software tasks
\citep{tian2024scicode,hu2026scicodeverified,hang2026pdeagentbench,anand2026mdarena,duston2025ainsteinbench}.
To improve its program, an agent needs a practical way to check how well the
current version meets those requirements.

Consider a task that prescribes the temperature at a domain boundary. After
each code revision, an agent needs to measure the current program against that
requirement. A prepared check makes the measurement repeatable. We ask whether
access to this ready-to-run implementation improves repair when both groups
retain the same written rule and ordinary Python execution.

Rules to Tools (R2T) packages public scientific requirements as callable
checking tools. The agent submits its current program to a check, reads the
measurement or error report, and continues editing. It can repeat this process
as the program changes. The tools cover measurable requirements such as boundary
values, physical relations, equations, and required output files. A separate
benchmark evaluator grades the finished program. The controlled comparison
measures access to the prepared implementation as a support package; retained
traces document how agents use its measurements.

The positive task-level effects occur on one initially flagged task and two
tasks whose initial checks report no violation. R2T evaluates ready access to
executable scientific checks during
revision; the final grader measures complete repair across the whole task.

Tool-based critique, executable specifications, and callable scientific
knowledge already support agent workflows
\citep{gou2023critic,akhond2025codemetaagent,liang2026sectdd,miao2026paper2agent}.
Our matched SciCode comparison holds the public scientific relation, probe,
tolerance, and ordinary execution access fixed while supplying one group with
a prepared check of the evolving program. Independent graders and separate
token and CPU records measure complete repair, component progress, and cost.

We study access to a prepared executable check through matched SciCode repair
comparisons and complementary experiments in four scientific computing
benchmarks. The central comparison uses eight task IDs absent from earlier
checker-development bindings and seven further task IDs. Task-specific written
definitions and executable checks are frozen before matched repair; an
independent evaluator scores each final program. Additional arms examine an
initial report and checker source delivered through ordinary Python.

The two task-ID cohorts yield 13/16 versus 15/16 and 13/14 versus 14/14.
Development-exposed SciCode programs test alternate starting artifacts;
matched PDE tasks measure completion and cost. Selected flow tasks and
historical molecular and repository panels retain their separate support
designs. Section~\ref{sec:design} specifies their assignments and scoring.

We make three contributions. First, R2T packages written public scientific
requirements as author-constructed callable checks of the program under
revision. Second, a
matched SciCode repair design holds written criteria, starting programs, model,
and budgets fixed while varying access to a prepared implementation; additional
arms examine how that implementation is delivered. Third, we measure complete
repair, native-step progress, and resource cost in the SciCode comparisons and
complementary studies across four scientific computing benchmarks, revealing
task-dependent gains and computational tradeoffs.

\section{Related Work}

\paragraph{Feedback-driven language agents.}
Self-Refine iteratively generates feedback and revises an answer with the same
language model \citep{madaan2023selfrefine}. Reflexion uses feedback and stored
verbal reflections to guide subsequent trials \citep{shinn2023reflexion}.
CRITIC obtains external tool feedback to support self-correction
\citep{gou2023critic}. Scientific auto-research agents also revise executable
training code using external evaluator feedback
\citep{ning2026specialistresearch}. Agent-written tests on SWE-bench Verified often supply
observations through printed values during repair \citep{chen2026rethinkingtests}.
These approaches establish feedback-driven revision as
a general agent pattern. R2T studies access to prepared checks within that pattern,
using public scientific requirements to construct measurements of the current
program and tracking both numerical progress and execution cost.

\paragraph{Scientific coding agents.}
SciCode and SciCode-Verified evaluate research-oriented numerical programming
\citep{tian2024scicode,hu2026scicodeverified}. PDEAgentBench evaluates solver
generation across equation families and libraries \citep{hang2026pdeagentbench}.
MDArena and AInsteinBench extend evaluation to molecular simulation workflows
and scientific software repositories \citep{anand2026mdarena,duston2025ainsteinbench}.
SWE-bench Science examines repository repair and ablates scientific guidance
\citep{xu2026swescience}, connecting task success to the content of that guidance.
Auto Research for Materials studies agent-written scientific learning workflows
and evaluates selected changes on held-out tasks
\citep{ning2026materialsresearch}.
Together these studies cover complete repairs and partial progress.

\paragraph{Scientific feedback.}
CodePDE combines partial differential equation (PDE) solver generation with
debugging and numerical refinement \citep{li2025codepde}. Lang-PINN uses symbolic
checks and feedback for physics-informed neural networks, which incorporate
equation constraints during training \citep{he2025langpinn}. These methods establish the utility of
feedback in scientific code generation. Our controlled comparisons examine the
contribution of executable public checks when agents already have complete
problem statements and ordinary execution access.

\paragraph{Executable specifications.}
CodeMetaAgent constructs tests through metamorphic specifications, and SecTDD
studies tests supplied before generation and feedback during repair using
identical initial programs and separate hidden-test evaluation
\citep{akhond2025codemetaagent,liang2026sectdd}. A metamorphic specification
describes how outputs should relate when inputs undergo a known transformation.
Scientific methodology checking also uses LLM-generated bug patterns
\citep{samsonau2026scicodelint}. CodeSpecBench evaluates behavioral
specification generation, and SpecCoder validates intermediate assertions on
correct executions and behavior-changing mutants
\citep{chen2026codespecbench,leanh2026speccoder}. CodeSpec compares textual and
executable architecture and behavior specifications during repository feature
development \citep{wang2026codespec}. R2T studies public numerical relations
as checks on evolving scientific programs through matched repair, independent
grading, and resource measurement.

\paragraph{Scientific knowledge as callable tools.}
Paper2Agent converts papers and code into callable scientific agents and
compares MCP tools with Markdown skills \citep{miao2026paper2agent}. Its
comparison studies how scientific knowledge reaches an agent through text and
execution. R2T applies executable checks to an evolving program, using public
requirements for measurements and independent graders for final artifacts.
Executable Code Knowledge links code with contracts and validation state
\citep{gao2026eck}. R2T measures repair and cost from prepared numerical checks.

\section{Executable Scientific Checks}
\label{sec:method}

\paragraph{A public requirement becomes a measurement.}
An \emph{artifact} is the code, output file, or numerical field being revised.
Let $P_t$ denote the complete public specification of task $t$ and let $c$ denote
the current candidate artifact. A diagnostic is a map
\begin{equation}
 d_j(c;P_t)=(\text{relation},\text{probe},\text{observation},\text{status}).
\end{equation}
The relation identifies a public requirement. The probe supplies its input
configuration. A \emph{witness input} is the configuration associated with a
reported discrepancy. The observation records a discrepancy or execution error.
The status distinguishes a discrete verdict from a continuous measurement or
an unavailable observation. Written rules and code implement this specification.

\paragraph{Specification equivalence and generated feedback.}
Let $S_t$ collect the public checking relations, probe inputs, tolerances,
applicability conditions, and reporting rules. The written representation $W_t$
describes $S_t$, and the tool $T_t$ implements it. We use
\emph{specification equivalence} for this correspondence of scientific checking
content. It applies at the level of the declared criteria and probes.
For a candidate $c$, the feedback is generated by execution,
\begin{equation}
 W_t=\operatorname{Describe}(S_t),\qquad
 T_t=\operatorname{Implement}(S_t),\qquad
 y_t=T_t(c).
\end{equation}
The operations denote engineered representations of a specification. The
task-ID-heldout and frozen alternate SciCode comparisons use this correspondence
at the declared criterion and probe level. A
\emph{residual} measures how far a candidate violates an equation. An error
location, residual, or convergence status in $y_t$ depends on the actual candidate
and becomes available through the computation. A text-supported agent can obtain
such observations by implementing and executing the corresponding procedure.
The supplied tool makes that procedure immediately callable. Independent
evaluators retain hidden solutions and final scores.

\paragraph{Benchmark bindings.}
An \emph{artifact contract} specifies required files, interfaces, or output
structure. A \emph{benchmark binding} identifies the files, entry points, and
numerical quantities used by a check. Reusable diagnostic patterns need concrete
bindings for public interfaces,
parameter values, output paths, and applicable physical regimes.
Table~\ref{tab:benchmark-map} shows these bindings across scientific artifacts.

Table~\ref{tab:public-check-map} follows four public requirements into
measurements of a current program. Each row starts with a relation in the
public task. Its probe and admitted error define the measurement. The returned
value describes the candidate's behavior. The SciCode text prompts include
the listed relations, inputs, and status policies. The flow rule card states a
pointwise field-capture gate; its implementation uses a grid-relative $L_2$
gate. Both use the public flow relations. The rule cards and implementations
are included under \texttt{supplement/public\_checker\_sources/} in the
anonymous supplement.

\begin{table}[!htbp]
\centering\footnotesize
\caption{\textbf{Public rules mapped to measurements of the current program.}}
\label{tab:public-check-map}
\setlength{\tabcolsep}{4pt}
\renewcommand{\arraystretch}{1.18}
\rowcolors{2}{white}{TableTint!30}
\begin{tabularx}{\linewidth}{>{\raggedright\arraybackslash}p{.22\linewidth}>{\raggedright\arraybackslash}X>{\raggedright\arraybackslash}p{.18\linewidth}>{\raggedright\arraybackslash}p{.20\linewidth}}
\toprule
\rowcolor{TableHead}
\textbf{Public source and rule} & \textbf{Probe on the candidate} & \textbf{Admitted error} & \textbf{Returned observation} \\
\midrule
SciCode 12 public radial equation and Numerov recurrence & $(E,\ell)=(-0.63,0)$, $(0.37,2)$, and $(-0.21,1)$ on grids of 65, 129, and 97 points & Relative residual $\leq 10^{-8}$ & Largest residual and its value in each setting \\
SciCode 22 public harmonic rotation and field reconstruction & Degrees 1 and 2; rotation of 0.41 radians about an axis proportional to $(1,2,-1)$; four directions & Relative discrepancy $\leq 10^{-7}$ & Largest field discrepancy and values by degree \\
SciCode 73 public reciprocal-cell geometry & Cell lengths $(3,4,5)$, angles $(88,96,104)^\circ$, polar maximum $31^\circ$, wavelength $1.2$ & Continuous discrepancy; no binary threshold & Worst reflection, ring key, expected spacing, and discrepancy \\
PDE flow public boundary and flow equations & Capture one complete vector field whose magnitude matches the returned grid; differentiate on the physical grid with four boundary layers cropped & Text uses pointwise relative matching; tool uses grid-relative $L_2\leq 10^{-10}$ & Boundary error, divergence and momentum residuals, or unavailable status \\
\bottomrule
\end{tabularx}
\end{table}

\paragraph{Scientific measurements.}
For a scalar PDE grid, an equation check computes
\begin{equation}
 r_h(u)=\frac{\|L_hu-f\|_2}{\|L_hu\|_2+\|f\|_2+10^{-12}},
\end{equation}
using physical grid spacing and an interior crop. Grid differentiation and
interpolation introduce numerical error, so this check returns a continuous
observation. Boundary discrepancies retain their measured magnitude. In coupled
flow, passive observers recover a vector field whose norm matches the returned
velocity-magnitude grid. The tool measures divergence and the curl of the
momentum equation, eliminating pressure gradients. Ambiguous captures receive
an unavailable status.

SciCode checks include a discrete radial recurrence, boundary-derivative
orientation, field reconstruction after rotation, a Dyson-equation residual,
converged quadrature of a public response definition, and reciprocal-cell
geometry. These examples illustrate how public scientific knowledge supplies
answer-free tests. Appendix~\ref{app:public-definitions}
gives their constructions, and Appendix~\ref{app:pde-delivery} specifies the
flow measurements.

\paragraph{Qualification and access.}
R2T uses task-specific checks authored or reviewed by the study team. The
eight-ID cohort used static preflight and execution on each starting module.
The seven-task extension additionally qualified checks on accepted programs
and deliberately constructed faults. Public-task generation produced 21 drafts;
public-law review and
accepted-program validation yielded 14 final checks. They passed 28/28
accepted-program evaluations and 8/8 additional evaluations on programs kept
outside selection, and detected 7/7 targeted seeded faults. One of seven
failed starting programs triggered a selected check. The interaction contains
public inputs, candidate code, and diagnostics; independent grading supplies
final scores. Appendix~\ref{app:extension-mechanism} records construction.

\paragraph{Artifact-level feedback.}
The agent inspects a candidate, obtains observations, saves a revision, and
checks the saved artifact again. Text-supported agents implement their probes
with ordinary Python. Tool-supported agents can call the supplied checker.
Failed checks permit revision, and budget exhaustion submits the current saved
artifact. This workflow makes both diagnostic coverage and final delivery
observable in the retained trajectory.

\section{Evaluation Design}
\label{sec:design}

\paragraph{Groups and information.}
We use the names \emph{text group} and \emph{tool group} throughout. The common
scientific content is specified at the requirement level. The matched SciCode
groups share written checks, probes, tolerances, applicability conditions, and
ordinary code execution; the treatment supplies a prebuilt implementation
through a dedicated command. The primary contrast measures this combined delivery.
Table~\ref{tab:benchmark-map} maps each benchmark to its artifact and support
design. The original SciCode and PDE studies add tools to shared definitions;
flow uses separate rule-card and implementation delivery with different
field-capture gates; historical MDArena
and AInstein retain written, checker-only, and combined conditions.
Specification equivalence applies to declared checks in the matched SciCode
studies. Table~\ref{tab:core-outcomes} names each outcome contrast.

\begin{table}[!htbp]
\centering\scriptsize
\caption{\textbf{Scientific artifacts and support designs across four benchmarks.}}
\label{tab:benchmark-map}
\setlength{\tabcolsep}{3pt}
\renewcommand{\arraystretch}{1.08}
\rowcolors{2}{white}{TableTint!30}
\begin{tabularx}{\linewidth}{>{\bfseries}p{.15\linewidth}>{\raggedright\arraybackslash}p{.17\linewidth}>{\raggedright\arraybackslash}X>{\raggedright\arraybackslash}p{.25\linewidth}}
\toprule
\rowcolor{TableHead}
Benchmark & Artifact & Public measurement & Support design \\
\midrule
SciCode & Numerical module & Boundary, recurrence, rotation, response & Written rules and prepared checks \\
PDE & Solver field & Boundary and equation residual & Matched checks; separate flow delivery \\
MDArena & Molecular workflow & Structure, protocol, artifact & Historical support forms \\
AInstein & Repository & Public API and domain probes & Historical support forms \\
\bottomrule
\end{tabularx}
\end{table}

\paragraph{Tasks and assignments.}
The original programming panels cover twelve SciCode and twelve PDE tasks;
historical panels cover eighteen MDArena tasks and fifteen AInstein rule units
from five repositories. The first task-ID SciCode cohort contains eight
archived failed starts, two continuations per group, and IDs absent from prior
checker bindings. Authors constructed the task-specific checks after archived
starts and context records existed, then froze the pair assignments and checks
before repair. Appendix~\ref{app:heldout-scicode} details these exposures.

A disjoint extension uses all seven remaining archived failed test-split task
IDs without prior checker bindings. Public-only generation proposes candidate
checks; review and accepted-program controls freeze two checks per task before
paired repair. Two additional access conditions on the first eight IDs supply
either an initial checker report or the frozen checker source for ordinary
Python execution. Appendix~\ref{app:extension-mechanism} reports construction,
qualification, and all assigned outcomes.

Five eligible alternate SciCode failures receive two trajectories per group
with a frozen enhanced checker and shared expanded written definitions. Their
different starting programs test sensitivity to program realizations
\citep{ning2026implementationlottery}. A seven-task flow cohort receives four
trajectories per group after a frozen textual development screen, detailed in
Appendix~\ref{app:pde-delivery}.

\paragraph{Execution and budgets.}
Controlled programming runs use DeepSeek Flash, three saved revisions,
196,608 reported output tokens, and 360 CPU-budget seconds per trajectory.
Each public execution has a 120-second wall cap and 3 GB memory cap; the
primary response cap is twenty-four. A separately reported flow sensitivity
allows one further response for eligible unfinished trajectories, carrying
forward consumed token, CPU, and revision budgets. Appendix~\ref{app:turn-display}
specifies both response caps and eligibility.

Historical workflow and repository experiments retain benchmark-native time
limits. Appendix~\ref{app:extended-tool-group} records a separate tool-only
evaluation of failed historical units.

\paragraph{Scoring.}
Complete-task success requires the native endpoint. SciCode requires every
scored subtask to pass. Both task-ID cohorts grade a final complete module on
the 2024 SciCode-Verified tests and retry only failed steps in the pinned 2025
environment. Each step passes if either environment accepts it, reducing
environment-specific failures under one shared endpoint. The twelve-task
shared-definition and development-exposed cohorts use the 2024 environment
alone. All these cohorts evaluate complete-module repair; the canonical
leaderboard uses stepwise generation. PDE requires relative grid error
at most $10^{-3}$ and
solve-call time at most 30 seconds. Historical workflows use their original
strict outcomes. Native partial scores use \emph{task-macro averaging}, which
averages repeated units within each task and then weights tasks equally. MDArena retains weighted
criterion values; AInstein retains parsed test completion; SciCode retains
scored-subtask completion. These measures have benchmark-specific meanings.

For operational score summaries, a timeout or unavailable valid evaluation
receives zero. The accompanying records flag unavailable test counts so that
this assignment has an explicit meaning. PDE reports numerical error and
runtime separately. Public diagnostic observations supply feedback and receive
no final-score credit. Output-token and CPU totals include failed attempts.
Recorded missing usage remains explicit in cost comparisons.

\section{Results}

We first report the matched task-ID-heldout SciCode comparison. The remaining
rows place it alongside earlier matched, selected, and historical support
designs. Native partial scores describe progress within unfinished programs,
and resource measurements account for the cost of obtaining that progress.

\begin{table}[!htbp]
\centering\footnotesize
\caption{\textbf{Complete-task outcomes by comparison design.} Rows within
each block retain their own independent task units and support forms.}
\label{tab:core-outcomes}
\setlength{\tabcolsep}{4pt}
\renewcommand{\arraystretch}{1.16}
\rowcolors{2}{white}{TableTint!30}
\begin{tabularx}{\linewidth}{>{\bfseries}lXcrr}
\toprule
\rowcolor{TableHead}
Benchmark & \textbf{Comparison} & \textbf{Tasks} & \textbf{Text} & \textbf{Tools} \\
\midrule
\multicolumn{5}{l}{\emph{Matched written definitions and prepared checks}} \\
SciCode & Task-ID-heldout matched checks & 8 & 13/16 & 15/16 \\
SciCode & Independent task-ID extension & 7 & 13/14 & 14/14 \\
SciCode & Shared-definition cohort & 12 & 13/24 & 13/24 \\
SciCode & Development-exposed alternate programs & 5 & 3/10 & 7/10 \\
PDE & Detailed shared definitions & 12 & 23/24 & 24/24 \\
\midrule
\multicolumn{5}{l}{\emph{Separate delivery-form and historical comparisons}} \\
PDE flow & 24-response protocol & 7 & 16/28 & 21/28 \\
PDE flow & 25-response sensitivity & 7 & 20/28 & 24/28 \\
PDE & Brief guidance vs. rules plus checks & 12 & 24/24 & 24/24 \\
MDArena & Written rules vs. rules plus checks & 18 & 5/18 & 7/18 \\
AInstein & Written rules vs. rules plus checks & 5 & 4/15 & 7/15 \\
\bottomrule
\end{tabularx}
\par\vspace{3pt}
\begin{minipage}{\linewidth}\footnotesize
\textit{Reading key.} Tasks count independent problems; AInstein contributes
three rule units per task. SciCode alternate programs reuse task IDs from
checker development. Flow changes rule-card delivery, access to an
implementation, and the field-capture gate; its 25-response row reuses the
24-response assignments. Brief-guidance PDE changes both written detail and
tool access. Historical rows use written rules and rules plus checks;
checker-only outcomes appear in Appendix~\ref{app:historical-panels}. The two
SciCode task-ID rows use the 2024/2025 per-step OR endpoint; the shared-definition
and development-exposed rows use the 2024 endpoint.
\end{minipage}
\end{table}

\begin{table}[!htbp]
\centering\footnotesize
\caption{\textbf{Access to a prepared check on the original eight SciCode IDs.}
Written rules and the dedicated command are original arms; initial report and
source through Python are added fresh arms. Native steps are task-macro means.
Model tokens include cache reads and output; the initial report's preparation
cost is outside its trajectory total.}
\label{tab:checker-access-main}
\setlength{\tabcolsep}{4pt}
\renewcommand{\arraystretch}{1.12}
\rowcolors{2}{white}{TableTint!30}
\begin{tabularx}{\linewidth}{>{\raggedright\arraybackslash}Xrrrr}
\toprule
\rowcolor{TableHead}
Access & Complete & Native steps & Model tokens & Public CPU \\
\midrule
Written rules & 13/16 & 95.83\% & 7.25M & 456.90s \\
Initial report & 14/16 & 98.96\% & 8.68M & 521.24s \\
Source through Python & 15/16 & 99.48\% & 8.51M & 509.39s \\
Dedicated command & 15/16 & 95.83\% & 7.88M & 633.15s \\
\bottomrule
\end{tabularx}
\end{table}

\paragraph{Held-out SciCode repair.}
All eight archived starting modules fail the complete-task endpoint. With the
checks and pair assignments frozen, the text group repairs 13/16 programs and
the tool group repairs 15/16. Averaging the two continuations within each task
gives a 12.5-point tool-minus-text difference. A \emph{task-cluster bootstrap}
resamples tasks with both repetitions; its 95\%
interval is $[-12.5,43.75]$ percentage points. Two tasks favor tools, one favors
text, and five tie. Task-macro native-step completion is 95.83\% in both groups.
The complete-repair gain therefore occurs alongside unchanged average
subtask completion in this cohort.

\begin{figure}[!htbp]
\centering
\includegraphics[width=\linewidth]{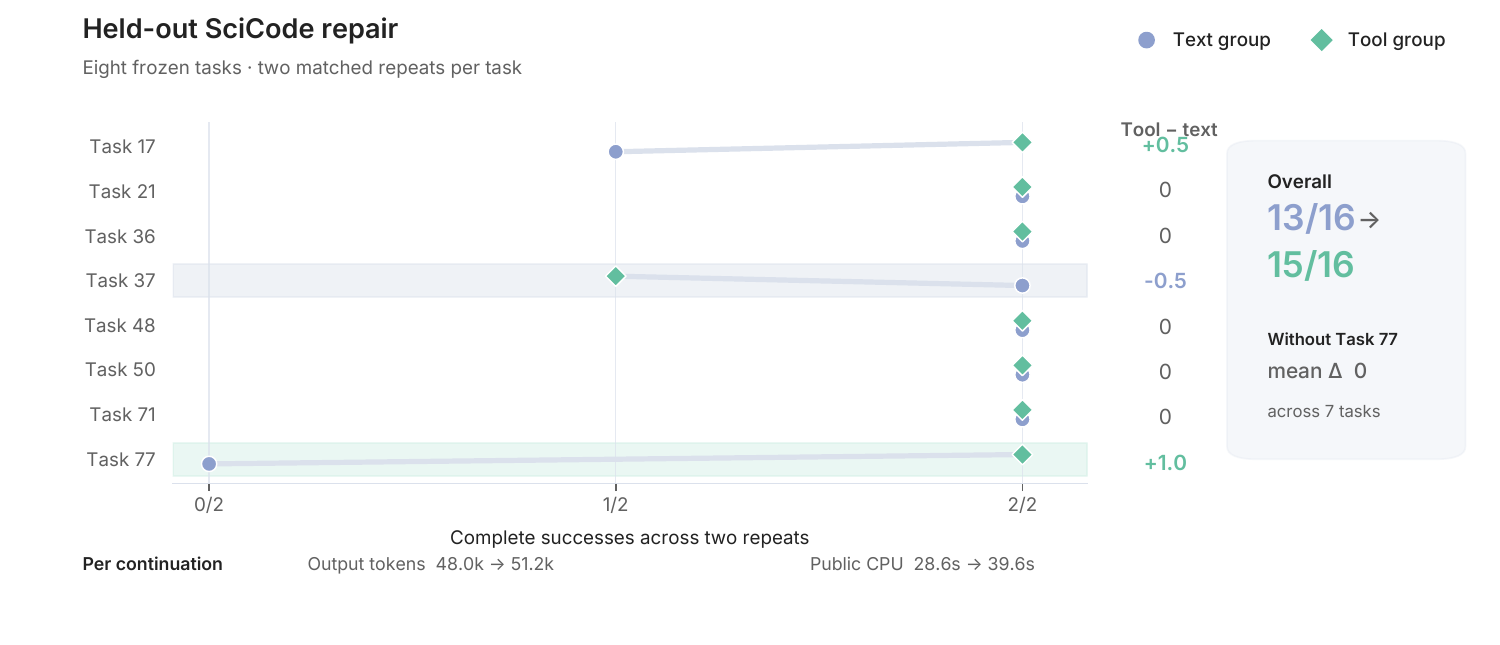}
\caption{Task-ID-heldout SciCode repair. Eight distinct task IDs contribute two
matched continuations per group. Complete-task success is 13/16 with text and
15/16 with tools. The task-macro tool-minus-text difference is 12.5 percentage
points and becomes zero when task 77 is excluded. Output-token and public-CPU
totals are reported separately in the text.}
\label{fig:scicode-heldout}
\end{figure}

On task 77, text repairs
0/2 programs and tools repair 2/2. Tasks 17 and 37 contribute one gain in
opposite directions. The net task-macro difference across the other seven
tasks is zero. Task 77 shares a molecular-dynamics family with checker-development
task 80. Figure~\ref{fig:scicode-heldout} shows the paired task outcomes,
and Appendix~\ref{app:heldout-scicode} gives the full protocol.

The independent seven-ID extension adds one tool-only repair on task 11
(1/2 to 2/2); tasks 27, 32, 34, 39, 55, and 62 tie at 2/2 in both arms.
Appendix~\ref{app:extension-mechanism} gives the full task table and
task-cluster uncertainty. Its bootstrap interval $[0,21.43]$ points has a zero
lower endpoint because no task favors text. Exact two-sided sign tests on
non-tied task effects give $p=1$ for the eight-ID (2:1) and seven-ID (1:0)
cohorts, and $p=0.625$ for their combined 3:1 pattern.

\paragraph{Additional SciCode support designs.}
The twelve-task shared-definition cohort ties at 13/24. A development-exposed
alternate-program comparison reaches 3/10 with text and 7/10 with checks.

\paragraph{Development-exposed SciCode records partial progress.}
On alternate archived failures, the frozen tool increases task-macro native
subtask completion from 61.27\% to 90.48\%. Three task-level success contrasts
favor tools and two tie. Task 80 remains incomplete in both groups, yet its
text trajectories pass 6/7 and 5/7 subtasks while both tool trajectories pass
6/7. The native partial measure preserves this progress. The comparison uses
five development-exposed tasks with different starting artifacts and matched written
definitions. Appendix~\ref{app:alternate-artifacts} retains every task result.

\begin{figure}[!htbp]
\centering
\includegraphics[width=\linewidth]{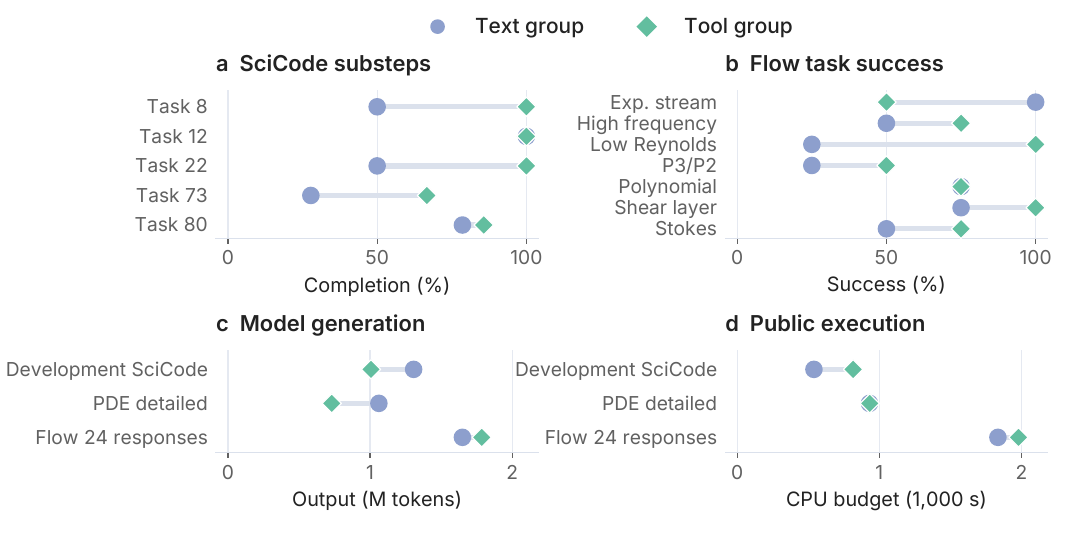}
\caption{Task-level progress and agent-side resource use in separate cohorts.
Panel a shows development-exposed SciCode alternate-program substeps; panel b
shows flow success under the primary 24-response cap. Panels c and d report
model output and public CPU for development-exposed SciCode, detailed-text PDE,
and 24-response flow. Main task-ID SciCode costs appear in the text.}
\label{fig:results-detail}
\end{figure}

\paragraph{Selected flow cohort.}
The flow cohort follows a frozen screen of 29 tasks: three initial programs
pass, ten return inaccurate fields, and sixteen fail execution. A textual
development probe leaves seven graded failures for paired comparison. Selection
on text-probe failure favors the tool group in this selected comparison.
The flow waves complete 16/28 text and 21/28 tool trajectories under the
24-response cap; task signs favor tools on five, text on one, and tie on one.
A secondary 25-response sensitivity gives 20/28 and 24/28, with signs 4:2:1.
The scalar-PDE pilot completes 4/4 in both groups. This selected cohort
compares shared public flow relations under different delivery forms and
capture gates
(Figure~\ref{fig:results-detail}; Appendix~\ref{app:pde-delivery}).

\paragraph{Checker access and observed revision traces.}
Table~\ref{tab:checker-access-main} separates complete repairs, native-step
progress, and resource use across the four access forms. The added
source-through-Python arm produces 40 completed public checker reports.
The initial-report and source-through-Python arms contain fresh continuations.
Task 77 yields 0/2 with the initial report, 1/2 with source through Python,
and 2/2 with the original dedicated-tool group. With two continuations per
arm, these task-level counts do not rank the access forms.

Matched outcomes measure the complete support package. Saved traces document
how agents used its measurements.
All sixteen held-out tool trajectories called the checker, making 34 calls;
initial reports flagged tasks 17 and 48. Task 77's initial checks reported no
violation, as did task 11's two initial checks in the seven-ID extension.
Both successful task 77 tool trajectories later repaired periodic wrapping
and pressure through ordinary Python. Development-exposed task 22 also begins
with all supplied checks reporting no violation and repairs code outside their
coverage. Task 17 has a positive effect with a flagged initial violation.
The task 77 and 11 gains arise without an initial checker-flagged error; these
traces document the full executable-support package. All ten
development-exposed tool trajectories called the checker, making 24 calls;
Appendix~\ref{app:alternate-artifacts}
includes an illustrative task 12 trajectory that both groups complete. Under
the original 24-response cap in coupled flow, 18 of 21
successful tool trajectories reduced a measured discrepancy after an edit.
The independent evaluator grades final success.

\paragraph{Generation and execution have different costs.}
In the eight-ID SciCode cohort, text-to-tool totals are 7.25M to 7.88M model
tokens ($+8.7\%$), 767,871 to 819,981 output tokens ($+6.8\%$), and 456.90
to 633.15 public CPU seconds ($+38.6\%$). The seven-ID cohort uses 5.78M to
4.52M total tokens and 322.53 to 826.73 CPU seconds. Across both cohorts,
reported output is at least 1,327,295 text tokens versus 1,262,059 tool
tokens, giving at least 4.9\% lower tool output; one seven-ID text request
has no returned usage. Public CPU rises from 779.43 to 1,459.88 seconds
($+87.3\%$).

Development-exposed SciCode saves 22.9\% reported output, with a 17.9\%
issued-output bound. Matched detailed-text PDE uses 1,062,099 versus 730,723
output tokens, a 31.2\% reduction and 25.0\% bound. Under the primary
24-response flow cap, text uses 1,649,905 output tokens and tools use
1,786,229; every request has recorded usage. The 25-response flow sensitivity
has two text requests without returned usage.

\section{Conclusion}

Rules to Tools supplies executable checks of public scientific requirements
during code revision. Across two SciCode task-ID cohorts, complete repair is
26/30 with text and 29/30 with checks. Tool-favoring tasks 17, 77, and 11
include one initial violation and two initial no-violation reports; task 37
favors text. Fresh source-through-Python and original
dedicated-command arms each score 15/16 on the first eight IDs. The twelve-task
shared-definition cohort ties at 13/24, while five development-exposed
alternate programs score 3/10 versus 7/10. Matched PDE checks score 24/24
versus 23/24 with detailed text and use 31.2\% fewer reported output tokens.
The selected flow comparison changes delivery form and capture gate.
Complete-repair effects and agent-side costs vary by cohort. Checker authoring
time was not recorded.

\section*{Reproducibility Statement}
Controlled SciCode/PDE/flow runs used DeepSeek-V4.1-Flash
(\texttt{deepseek-flash}) on 17 and 24--25 September 2026 UTC with temperature
1, top-$p$ 0.95, and high-effort thinking. Historical MDArena/AInstein runs
used \texttt{deepseek-v4-flash} in August; extra checker-only attempts used
\texttt{deepseek-flash} in September. Anonymous \texttt{supplement.zip}
contains per-phase settings, SciCode task inputs and prompts, public rules,
and checker sources.

\section*{AI Use Statement}
AI assisted literature retrieval, experiment design and execution, analysis,
visualization, and drafting. DeepSeek generated evaluated agent trajectories;
independent benchmark evaluators supplied final scores. Authors retain final
review responsibility.

\bibliographystyle{iclr2027_conference}
\bibliography{scientific_scope_references,references}

\appendix
\section{Response Caps and Sensitivity}
\label{app:turn-display}

The 24-response runner advances its consumed-response counter after preparing
tool replies. Its final reply therefore displays one response remaining at the
24-response stopping point. Both conditions receive this display, and the
actual primary comparison cap is 24 responses. A secondary 25-response
sensitivity, defined after those trajectories, continues eligible unfinished
states for one response. It carries forward token, CPU, and edit consumption;
eligibility uses interaction state and resource budgets, with final grades
outside the agent. The 24- and 25-response outcomes retain separate labels.

\section{Public-Definition Diagnostic Construction}
\label{app:public-definitions}

\paragraph{Discrete radial equation and boundary orientation.}
For the public equation $u''=fu$ on a uniform grid with spacing $h$, the Numerov
diagnostic computes
\begin{align}
 a_i &= (1-h^2 f_{i+1}/12)u_{i+1}, &
 b_i &= 2(1+5h^2 f_i/12)u_i,\\
 c_i &= (1-h^2 f_{i-1}/12)u_{i-1}, &
 r_N &= \frac{\|a-b+c\|_2}{\|a\|_2+\|b\|_2+\|c\|_2+10^{-12}}.
\end{align}
The public Rydberg convention gives
$f=\ell(\ell+1)/r^2-2/r-E$ for hydrogen and
$f=\ell(\ell+1)/r^2+(U_H-2Z)/r-E$ for the Hartree variant.
The coefficient is constructed from the public equation independently of the
candidate's coefficient routine. The diagnostic probes three energy and angular
momentum configurations on grids of 65, 129, and 97 points and admits a residual
tolerance of $10^{-8}$. A separate check inspects the sign of the outer derivative
in increasing physical radius. The prescribed negative derivative remains
negative after positive normalization. This sign is part of the requested
wavefunction convention; squared-density observables can conceal its reversal.

\paragraph{Field reconstruction under rotation.}
The public convention defines a spherical-harmonic transformation through
\begin{equation}
 Y_{\ell m}(r)=\sum_{v=-\ell}^{\ell}T_{\ell vm}(Q)Y_{\ell v}(Qr).
\end{equation}
The tool evaluates the basis independently and asks the candidate for the
transformation coefficients. It probes degrees 1 and 2, a rotation of 0.41 radians
around the normalized axis $(1,2,-1)$, and four generic spatial directions.
Qualification uses an independent Cartesian polynomial construction at these
degrees. A deliberately transposed rotation supplies a negative control. It
preserves the original identity, inverse, and norm checks while violating field
reconstruction. This is a synthetic checker test, with no contribution to the
benchmark repair count. Higher-degree recurrence correctness remains outside
this particular measurement.

\paragraph{Numerical integration of a density-response definition.}
For a spin-degenerate parabolic band at zero temperature, the public integral
can be evaluated after shifting the first occupied disk. With
$\alpha=E_F/k_F^2$ and $z=\omega+i\gamma$, the implemented numerical reference is
\begin{equation}
 D(q,z)=\frac{1}{2\pi^2}\int_0^{k_F}p\,dp\int_0^{2\pi}d\theta
 \left[\frac{1}{\alpha(2pq\cos\theta-q^2)-z}
       -\frac{1}{\alpha(2pq\cos\theta+q^2)-z}\right].
\end{equation}
Radial Gauss--Legendre quadrature of orders 32 and 64 is combined with four times
as many uniform angular points. The tool increases the radial order to 128 when
the estimates' relative difference exceeds the $10^{-8}$ quadrature refinement
threshold. An unresolved integral produces a diagnostic error. Candidate outputs
use a $10^{-6}$ discrepancy tolerance against the numerical reference. Smooth
complex frequencies avoid singular real-axis quadrature.

The probes enforce $n=k_F^2/(2\pi)$ and $v_F=2E_F/k_F$ with $E_F=18$ and $k_F=0.23$,
using the two configurations $(q,\omega,\gamma)=(0.06,9,6)$ and $(0.23,11,12)$.
This measures agreement on physically consistent parameter combinations.
Independent argument combinations and other frequency regimes retain separate
coverage requirements. Agents receive discrepancy and refinement measurements,
with the numerical reference values kept inside the checker. Two accepted
archived components qualify the numerical reference. A companion check tests
the public matrix equation $D=D_0I+D_0(V_qF)D$ with the stated unit conversion.

\paragraph{Reciprocal-cell geometry.}
For the direct-cell metric $G$ built from the public lattice lengths and angles,
each reflection $h$ has reciprocal magnitude $\sqrt{h^\top G^{-1}h}$. The tool
compares this quantity with the returned ring key. It reports the largest
discrepancy and its reflection, retaining rounding effects as observations.
Sorted reflection lists and unique ring membership are checked separately.
This construction uses public geometry and supplies no hidden ring list or
private grading tolerance.

\section{Task-ID-Heldout SciCode Repair}
\label{app:heldout-scicode}

The SciCode-Verified v2 source manifest lists 64 main problems. This repair
study uses eight test-split task IDs, 17, 21, 36, 37, 48, 50, 71, and 77,
that were absent from the earlier checker-development bindings. An archived
complete starting module for each ID failed the full-task endpoint in the
2024/2025 environment check. The eight IDs and two matched continuations per
group were fixed before the paired repair outcomes. Task-specific public checks
were constructed and validated before paired repair. Earlier context
experiments generated and graded the starting modules. Task 36 had a prior
numerical-error diagnosis, and task 77 shares a molecular-dynamics family
with development task 80.

One public rule source generates the written card and executable checker for
each task. The study team authored this task-specific source after archived
failed starting programs and earlier context records existed; no blinded
authoring log was retained. Its 25 rules across eight tasks specify a target function, witness
input, public relation, and tolerance. Both groups receive the same starting
module, public task, written definitions, model, and budgets. The text group
can run ordinary Python; the tool group can also call the prepared checker.
Public-only tool containers expose candidate code and diagnostic inputs. The
independent grader sees the finished module after the interaction closes. The
frozen public checks include a task 37 witness, a nonsingular task 48 probe,
and the task 77 displacement-sign convention. Static preflight verified these
definitions before paired repair outcomes were observed.
All eight task-bound checkers passed static qualification and executed on their starting
modules. This cohort has no uniform accepted-program and seeded-fault panel;
the seven-ID extension reports that additional qualification. Resource totals
cover agent repair trajectories; checker authoring time was not recorded.

Each final module is scored on all native substeps in the 2024 environment.
Only non-passing steps are retried in the 2025 environment. A task succeeds
when every substep passes in at least one environment. This endpoint evaluates
complete-module repair; the canonical leaderboard evaluates stepwise code
generation. All 32 assigned continuations completed and received final grades.

\begin{table}[!htbp]
\centering\footnotesize
\caption{\textbf{Task-ID-heldout SciCode full-module repair outcomes.} Each group has two matched continuations per task.}
\label{tab:heldout-scicode}
\setlength{\tabcolsep}{6pt}
\renewcommand{\arraystretch}{1.15}
\rowcolors{2}{white}{TableTint!30}
\begin{tabularx}{\linewidth}{>{\bfseries}p{.15\linewidth}>{\centering\arraybackslash}p{.23\linewidth}>{\centering\arraybackslash}p{.13\linewidth}>{\centering\arraybackslash}p{.13\linewidth}>{\raggedleft\arraybackslash}X}
\toprule
\rowcolor{TableHead}
Task ID & Initial violation & Text & Tools & Tool-minus-text \\
\midrule
17 & Yes & 1/2 & 2/2 & $+50$ pp \\
21 & No & 2/2 & 2/2 & $0$ pp \\
36 & No & 2/2 & 2/2 & $0$ pp \\
37 & No & 2/2 & 1/2 & $-50$ pp \\
48 & Yes & 2/2 & 2/2 & $0$ pp \\
50 & No & 2/2 & 2/2 & $0$ pp \\
71 & No & 2/2 & 2/2 & $0$ pp \\
77 & No & 0/2 & 2/2 & $+100$ pp \\
\midrule
All tasks & 2/8 & 13/16 & 15/16 & $+12.5$ pp \\
\bottomrule
\end{tabularx}
\end{table}

The task-macro tool-minus-text difference is 12.5 percentage points. A
10,000-draw task-cluster bootstrap with seed 20260923 yields a 95\% interval
of $[-12.5,43.75]$ points. Two tasks favor tools, one favors text, and five
tie. Removing task 77 gives a zero difference across the other seven tasks,
with interval $[-21.43,21.43]$ points. The task 77 starting program passes
all five supplied public checks. Its two successful tool trajectories later
use ordinary Python to revise periodic wrapping and pressure calculations.
The complete records retain prompts, checks, model continuations, grader
results, and output-token and public-execution accounting.

\section{Independent SciCode Extension and Checker Access}
\label{app:extension-mechanism}

\paragraph{Independent task IDs and frozen checks.}
The archive contains 25 distinct test-split task IDs with a saved failed
complete module under the 2024/2025 full-module endpoint. Eight IDs entered
the first task-ID-heldout cohort, and ten other IDs had earlier checker
development bindings. The seven remaining IDs, 11, 27, 32, 34, 39, 55, and
62, form the complete extension cohort. Each starts from the lowest-numbered
failed native replicate with every released code step saved. Initial private
grades confirm failure on all seven modules.

One fixed DeepSeek Flash request per task received the public problem, step
descriptions, function headers, and return contracts. Seven calls proposed
21 parseable checks. Public-law review and execution on accepted archived
programs selected two checks per task. The selected set includes three
manually specified public identities and one task-27 public function-signature
binding. Proposals, rule-review decisions, final written definitions,
executable checks, and the pair manifest were saved before paired repair. Generation requests
contained no candidate code. Hidden tests and final grades remained outside
the generation requests, executable checks, and repair prompts.

The 14 selected checks passed 28 evaluations on two accepted programs per
task. Four tasks had a third accepted program kept outside selection; its
eight check evaluations passed. Seven deliberately introduced faults, each
targeting one selected relation, triggered their intended checks. The seven
actual failed starting modules triggered a selected check in one case, task
34. This separates false-alarm qualification, targeted-fault detection, and
coverage of naturally occurring starting failures. Draft generation used
19,599 uncached input tokens, 11,559 output tokens, and 49.2 seconds of API
wall time. Manual public-law review has no timed record.

The two support groups share the full public task, the selected relations,
probe inputs, tolerances, starting module, model, and repair allowance. The
tool group can also call the frozen implementation. Each task receives two
fresh continuations per group and independent full-module grading. Table
\ref{tab:extension-outcomes} retains all assigned results.

\begin{table}[H]
\centering\footnotesize
\caption{\textbf{Seven independent SciCode task IDs.} The starting module
triggers a selected public check only for task 34. Each group receives two
continuations per task.}
\label{tab:extension-outcomes}
\setlength{\tabcolsep}{6pt}
\renewcommand{\arraystretch}{1.15}
\rowcolors{2}{white}{TableTint!30}
\begin{tabularx}{\linewidth}{>{\bfseries}p{.18\linewidth}>{\centering\arraybackslash}p{.26\linewidth}>{\centering\arraybackslash}p{.18\linewidth}>{\centering\arraybackslash}X}
\toprule
\rowcolor{TableHead}
Task ID & Initial check violation & Text & Tools \\
\midrule
11 & No & 1/2 & 2/2 \\
27 & No & 2/2 & 2/2 \\
32 & No & 2/2 & 2/2 \\
34 & Yes & 2/2 & 2/2 \\
39 & No & 2/2 & 2/2 \\
55 & No & 2/2 & 2/2 \\
62 & No & 2/2 & 2/2 \\
\midrule
All tasks & 1/7 & 13/14 & 14/14 \\
\bottomrule
\end{tabularx}
\end{table}

The task-macro repair difference is 7.14 percentage points. Task-cluster
resampling with 10,000 draws gives a 95\% interval of $[0,21.43]$ points.
Task 11 supplies the single difference in complete repairs. Its starting
module passes both selected public checks, and the successful tool trajectories
use checks around their revisions. This records a useful executable-support
package on one new ID without assigning the repair to an initially flagged
violation. Tools use 4,517,829 total model tokens, 442,078 output tokens,
826.73 public CPU seconds, and 2,879.9 summed interaction wall seconds. Text
uses 5,780,325 total tokens, 559,424 output tokens, 322.53 public CPU seconds,
and 3,350.1 recorded wall seconds. One interrupted text API request has no
returned token usage, so its token totals are lower bounds and the 21.8\%
model-token saving is a lower bound. The text wall record includes a 293.3-second
interrupted-request window; excluding that entire interval gives a 3,056.8-second
active-wall lower bound. Public execution consumes 504.20 additional CPU seconds
with tools.

\paragraph{Access ablations on the original eight IDs.}
Two further conditions keep the original failed starting modules, public
rules, model, budgets, and native endpoint. The \emph{initial-report} arm
receives the saved checker report on the starting module and uses ordinary
Python thereafter. Its initial report is supplied before the interaction,
with its preparation cost outside that arm's resource total. The
\emph{source-through-Python} arm receives the exact frozen checker source at
\texttt{/workspace/public\_checker.py} and runs it with ordinary Python. It
has no dedicated checker command. Both conditions receive two fresh
continuations for each of the eight IDs. Table~\ref{tab:checker-access-main}
reports the original and added arms together in the main paper.

\begin{table}[!htbp]
\centering\footnotesize
\caption{\textbf{Complete repairs by task and checker-access form.} Each cell
contains two continuations. Written-rule and dedicated-command outcomes are
the original matched arms; the other two columns contain fresh continuations.}
\label{tab:checker-access-tasks}
\setlength{\tabcolsep}{5pt}
\renewcommand{\arraystretch}{1.12}
\rowcolors{2}{white}{TableTint!30}
\begin{tabularx}{\linewidth}{>{\bfseries}p{.12\linewidth}>{\centering\arraybackslash}p{.17\linewidth}>{\centering\arraybackslash}p{.19\linewidth}>{\centering\arraybackslash}p{.22\linewidth}>{\centering\arraybackslash}X}
\toprule
\rowcolor{TableHead}
Task & Written & Initial report & Source in Python & Dedicated \\
\midrule
17 & 1/2 & 2/2 & 2/2 & 2/2 \\
21 & 2/2 & 2/2 & 2/2 & 2/2 \\
36 & 2/2 & 2/2 & 2/2 & 2/2 \\
37 & 2/2 & 2/2 & 2/2 & 1/2 \\
48 & 2/2 & 2/2 & 2/2 & 2/2 \\
50 & 2/2 & 2/2 & 2/2 & 2/2 \\
71 & 2/2 & 2/2 & 2/2 & 2/2 \\
77 & 0/2 & 0/2 & 1/2 & 2/2 \\
\midrule
All & 13/16 & 14/16 & 15/16 & 15/16 \\
\bottomrule
\end{tabularx}
\end{table}

The source-through-Python trajectories run the checker and produce 40
completed public reports. Task 77 has 0/2 complete repairs with written
rules, 0/2 with an initial report, 1/2 with source through Python, and 2/2
with the dedicated command. These two-repetition task rates cannot rank the
access forms; task 77 has no initial public-check violation. The dedicated
command and source-through-Python
conditions each reach 15/16 over all eight IDs, with different task-level
outcomes. Written rules and the dedicated command each average 95.83\%
native-step completion despite their different complete-repair counts.
The two endpoints capture different aspects of repair. Dedicated-command and
source-through-Python mean interaction wall times are 329.9 and 299.5 seconds
per trajectory; the initial-report and
written-rule conditions use 276.0 and 309.8 seconds, respectively. The
initial-report arm receives a precomputed measurement and therefore has a
different full-workflow cost boundary.

\section{Development-Exposed Alternate-Artifact Study}
\label{app:alternate-artifacts}

A supplementary frozen-tool experiment uses the available alternate archived
failures for tasks 8, 12, 22, 73, and 80. Enumeration covers the source family of every
original SciCode task. Ten of twelve task families have a complete, graded
alternate artifact. Alternate generations for tasks 67 and 69 stop at the
response limit before completing their code and receive no full-task grade.
Five of the ten graded alternates already pass; the five failed alternates
define the full supplementary repair cohort.
Each receives two trajectories per group, for twenty assignments. Both groups
receive the complete task and identical expanded written definitions matching
the enhanced diagnostics. The callable implementation and total budgets remain
frozen. The 13/24 shared-definition text cohort and this expanded-definition
text cohort are reported separately. The design tests
robustness to different initial programs on development-exposed tasks. All twenty final
grades are resolved. The per-task text/checker counts are 2/2 versus 2/2 on task
12, 0/2 versus 2/2 on task 22, 0/2 versus 1/2 on task 73, 1/2 versus 2/2 on task
8, and 0/2 versus 0/2 on task 80. The task-mean success difference is 40 percentage
points. Exploratory task-cluster resampling gives an interval of [10,70] points;
with only five task clusters, this describes the observed panel. An exact
two-sided sign test on the three non-tied task effects gives $p=0.25$.

The preparation code assembles each expanded written definition from a frozen
check's identifier, target function, witness input, public relation, and
diagnostic tolerance. Both groups receive that definition list in their
prompts. The tool group additionally receives the unchanged implementation
through \texttt{check\_current}. Numerical observations arise when a group
executes its checks on the current program. Prompt and runner files retain
the mapping.

The alternate task 22 artifact has no violation on its four supplied rotation
checks at the starting state or after revision. The
tool group repairs 2/2 complete tasks and the text group repairs 0/2. Both
successful tool trajectories revise translation coefficients beyond the
supplied rotation checks. The measured support package includes ordinary
investigation alongside no-violation check reports.

\begin{figure}[!htbp]
\centering
\includegraphics[width=\linewidth]{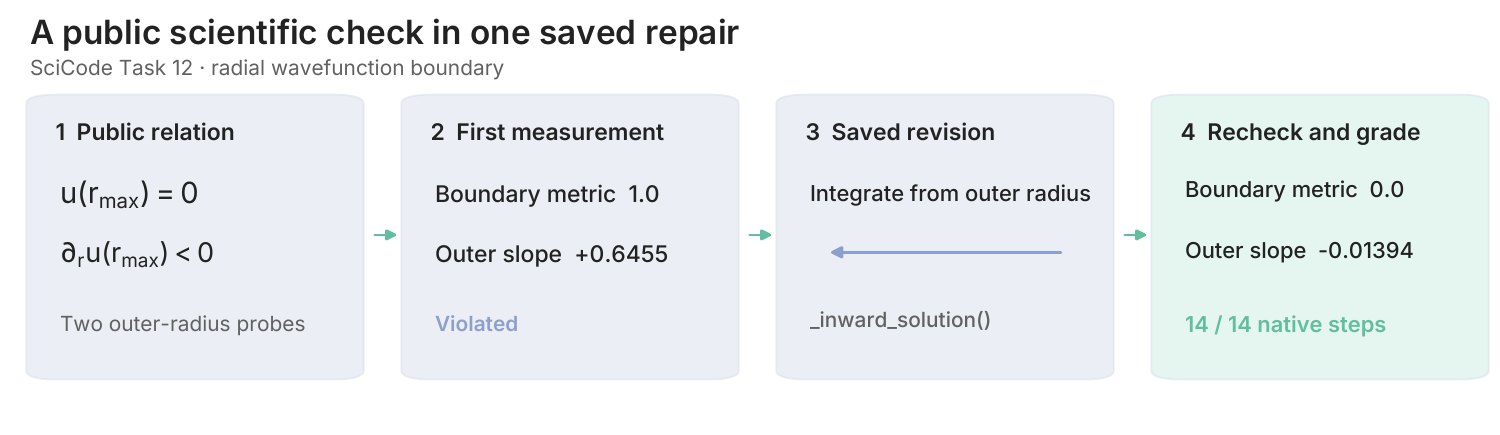}
\caption{An illustrative SciCode Task 12 trajectory. A public boundary check
reports a violation, followed by an inward-integration edit and consistent
rechecks. The final program passes all 14 native steps. Both text trajectories
also complete Task 12; this example documents checker use within repair.}
\label{fig:task12-trace}
\end{figure}

Across the ten trajectories per group, expanded written rules use 1,306,590
reported output tokens and frozen enhanced checks use 1,006,870, a 22.9\%
reduction. One checker-arm request has no returned usage and an issued output
cap of 65,536 tokens. Charging the full cap gives a reduction of at least
$1-(1{,}006{,}870+65{,}536)/1{,}306{,}590=17.9\%$.

\section{Historical Support-Form Panels}
\label{app:historical-panels}

The retained MDArena study contains eighteen task--rule units on eighteen tasks.
AInsteinBench contains fifteen units from three rules on each of five tasks.
Each historical unit has one execution per support form. Written rules recover
5/18 and 4/15 units; checker delivery recovers 8/18 and 7/15; rules plus checks
recover 7/18 and 7/15, respectively. The closest historical contrast to the new
matched-written-information experiment is written rules versus rules plus
checks. Historical checker-only support remains a separate condition.

The historical workflow includes automatic feedback delivery. Eight MDArena
configurations have recorded directory/file trigger conflicts; four still receive
other-stage feedback. Assigned outcomes remain in the reported counts, with
actual feedback exposure audited separately. Multiple rule units share each
AInstein task, and the new experiments repeat each task twice. Across the four
original panels there are forty-seven distinct tasks and fifty-seven task--rule units.
We retain separate benchmark strata and all three historical support forms.

Native partial scores provide the complementary view. For MDArena, mean reported
criterion scores are 70.81\%, 78.80\%, and 76.90\% for written rules, checker-only,
and their combination. For AInsteinBench, corresponding native partial diagnostics
are 98.73\%, 92.86\%, and 99.48\%. The checker-only AInstein value includes one
unparsed test log with a recorded zero fallback; its underlying partial-completion
bounds are [92.86\%,99.52\%]. The failed full-task outcome and missing test counts
remain visible. MDArena criteria include artifact checks, and AInstein uses
parsed test results. These benchmark-native quantities have separate meanings.

The companion data export contains every recorded SciCode native substep,
every MDArena criterion with its original weight and value, and each historical
task--rule unit's score and available parsed test counts. Repeated views of the
same unchanged trajectory retain their version label. Native substeps with
several assertions retain their native aggregate step score.

\section{Mutually Exclusive PDE Support Delivery}
\label{app:pde-delivery}

This additional study compares two support forms built from shared public flow
relations. The text group receives a rule card. The tool group receives a
callable implementation and minimal calling instructions, with the rule card
absent from its prompt. Both receive the complete public problem, the same
initial program, library guidance, ordinary Python execution, and matched
resource caps. The contrast combines delivery form, access to a prepared
implementation, and the field-capture gate specified below. The earlier
matched-written-information panels retain their rules-plus-tool treatment.

\paragraph{Screening and development.}
A frozen scalar-PDE screen includes eighty complete public cases, comprising twenty-seven
heat, fifteen reaction--diffusion, twenty-one Poisson, and seventeen
convection--diffusion problems. Seventy-four initial programs pass the retained
endpoint. One fails numerical accuracy, one exceeds the solve-time limit, and
four fail the execution interface. The two numerical or runtime failures form
the scalar delivery pilot. All eighty outcomes are retained.

A separate frozen screen covers twenty-nine complete unit-square Stokes and
Navier--Stokes cases. Three initial programs pass, ten return a field that fails
the reference-accuracy endpoint, and sixteen fail execution. The ten executable
failures receive a one-response textual development probe with no execution
interface. Two probe repairs pass, seven fail grading, and one response is
truncated. All seven graded failures enter the flow comparison, starting from
their original executable programs. Each receives two trajectories per group
and a second complete wave of two fresh trajectories per group. The second wave
copies the frozen initial prompts and checker source. Selection therefore
conditions the study on text-probe failure, favoring the tool arm in this
selected cohort. All subsequent assignments enter the comparison.

\paragraph{Flow diagnostics.}
The public output is velocity magnitude. Passive observers capture vector
samples from DOLFINx evaluation calls while preserving their arguments and
return values. Vector diagnostics require a unique complete captured field
whose magnitude matches the returned grid to relative tolerance $10^{-10}$.
Ambiguous or missing capture produces an unavailable result. Both groups may
implement passive observers in ordinary Python.

The rule card describes a pointwise relative magnitude match. The tool
implementation uses relative $L_2$ discrepancy across the output grid with the
same numerical threshold. These gates can admit different captured fields and
therefore change access to subsequent vector-equation feedback. Both forms use
the public flow relations and report those measurements after a unique complete
capture under their respective gates.

For a captured velocity field $v$, the instrument measures divergence and
\begin{equation}
  \nabla\!\times\!\left[-\nu\Delta v+(v\cdot\nabla)v-f\right],
\end{equation}
omitting convection for Stokes flow. Taking the curl removes the pressure
gradient. Spatial derivatives use physical grid spacing, second-order edge
formulas, and an interior crop of four grid layers. Reports include absolute
RMS residuals and normalized observations. Divergence uses the full velocity
gradient and velocity-over-domain-length scale. Momentum curl uses the sum of
component-curl norms and underlying momentum-component norms divided by domain
length; both denominators include $10^{-12}$. These observations retain
discretization error and carry no binary accuracy threshold. Boundary checks
use the supplied public Dirichlet data. Solver observers report available
convergence reasons and iteration counts.

Qualification includes accepted archived solvers, failed initial programs,
synthetic valid and invalid fields, and a discarded-field control. Accepted
artifact outputs remain unchanged under observation. Hidden reference arrays
and grading code stay outside the repair environment. Final evaluation retains
reference-field error and solve time as separate native measurements, alongside
joint success and unavailable-result counts. Every repetition is exported.
Diagnostics receive no final-score credit.

\begin{figure}[!htbp]
  \centering
  \includegraphics[width=\linewidth]{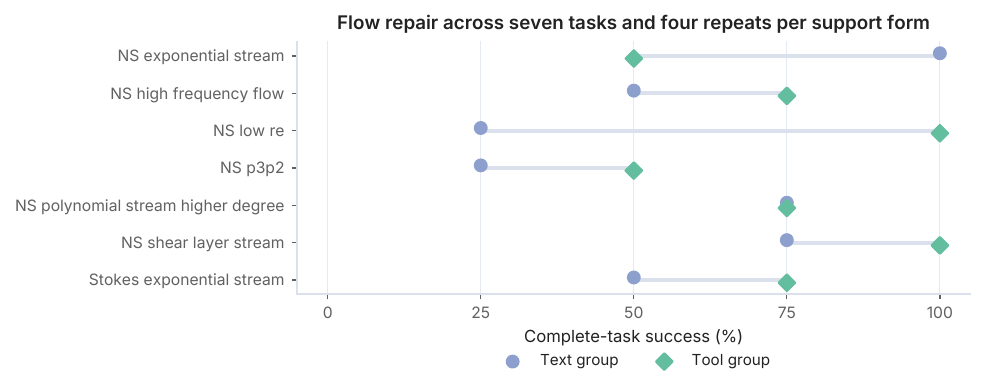}
  \caption{Mutually exclusive support delivery on the complete frozen flow
  cohort. Every task has four trajectories per support form across two waves.
  The displayed outcomes use a twenty-four-response cap. Appendix~\ref{app:turn-display}
  specifies the separately reported twenty-five-response sensitivity.
  NS denotes Navier--Stokes.}
  \label{fig:raw-flow-delivery}
\end{figure}

\paragraph{Complete raw outcomes.}
The scalar pilot recovers 4/4 assignments in both groups. Flow waves recover
9/14 and 7/14 with text, versus 11/14 and 10/14 with tools. Combined task-mean
success is 57.14\% versus 75.00\%, a 17.86-point difference. Five tasks favor
tools, one favors text, and one ties (Figure~\ref{fig:raw-flow-delivery}).
Reported output is 1,649,905 versus 1,786,229 tokens; public CPU is 1,834.37
versus 1,979.01 seconds, including failures.
The shared tasks and development screen define the measured population.

\paragraph{Response-cap sensitivity.}
A secondary twenty-five-response analysis continues unfinished states eligible
under the retained token, execution, and revision budgets. Eligibility uses
interaction state and excludes final grades. All thirty-eight eligible
continuations complete. Success is 20/28 (71.43\%) with text and 24/28
(85.71\%) with tools; task signs favor tools on four, text on two, and tie on
one. Per-trajectory outcomes remain in the data export.

Cumulative reported output is 1,851,667 text and 1,936,923 tool tokens. Two
text API requests have no returned usage; their combined issued-output cap is
131,072 tokens, spanning the reported difference. Charged execution is
1,867.72 versus 2,003.48 CPU-budget seconds. Tool CPU use is higher; the
output-cost ordering remains unresolved.

The data export retains each repetition's reference error, solve time, outcome,
and unavailable fields. In one text trajectory, isolated execution of the
boundary-expression path confirms two numerical Newton solves. This adherence
check supplies no additional repair outcome.

\clearpage
\section{PDE Generation Costs by Support Form}
\label{app:pde-brief-cost}

\paragraph{Matched detailed definitions.}
Detailed written rules and rules plus a prepared checker share the public
definitions across 24 repair trajectories. Complete repair is 23/24 and 24/24.
Reported output is 1,062,099 versus 730,723 tokens, a 31.2\% reduction with
checks. One checker request has no returned usage and an issued output cap of
65,536 tokens; charging that full cap gives a reduction of at least
$1-(730{,}723+65{,}536)/1{,}062{,}099=25.0\%$. Public CPU totals are 930.92
and 932.46 seconds.

\begin{figure}[H]
\centering
\includegraphics[width=.60\linewidth]{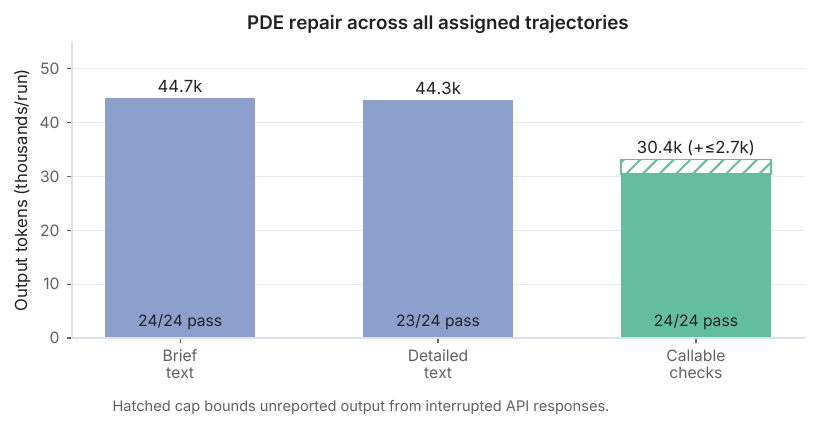}
\caption{PDE output per repair under brief text, detailed text, and callable
checks. Completion is 24/24, 23/24, and 24/24; hatching bounds one request
without returned checker usage.}
\label{fig:pde-checker-cost}
\end{figure}

\paragraph{Brief-guidance comparison.}
The separate brief-guidance text condition also completes 24/24, using
1,071,761 reported output tokens and 875.67 public CPU seconds. The checker
group retains its 730,723 reported tokens, 932.46 CPU seconds, and one
65,536-token missing-usage cap. Figure~\ref{fig:pde-checker-cost} shows
both text support forms against the same checker runs.

\paragraph{No-exact-solution PDE ceiling screen.}
A separate 29-task PDEAgentBench screen passed 29/29 initial native endpoints;
one matched revision pair tied, and 28 pairs were unstarted.

\section{Extended Tool-Group Evaluation}
\label{app:extended-tool-group}

Additional checker-only attempts cover ten MDArena and eight AInstein failed
units; seventeen finish and one stopped MDArena unit retains its grade.
Cumulative counts retain fifteen historical successes.

The attempts retain native evaluators, checker-call allowance, and time limits.
Two MDArena attempts use public artifact-trigger bindings; AInstein reuses
public probes. Task-macro partial scores assign zero to unavailable grades.

\begin{table}[H]
\centering\small
\caption{\textbf{Extended tool-group outcomes.}}
\label{tab:extended-tool-group}
\setlength{\tabcolsep}{6pt}
\renewcommand{\arraystretch}{1.3}
\rowcolors{2}{white}{TableTint!30}
\begin{tabularx}{\linewidth}{>{\bfseries}Xrrrr}
\toprule
\rowcolor{TableHead}
Benchmark & \textbf{Original} & \textbf{Extended} & \textbf{Native partial (\%)} & \textbf{Added output (k)} \\
\midrule
MDArena & 8/18 & 10/18 & 84.49 & 623.6 \\
AInsteinBench & 7/15 & 9/15 & 93.23 & 333.3 \\
\bottomrule
\end{tabularx}

\begin{minipage}{\linewidth}\footnotesize
\textit{Reading key.} Original refers to the historical checker-only arm;
Table~\ref{tab:core-outcomes} lists written rules plus checks. Added output
sums reported usage for new attempts. Task-macro partial scores give zero to
an unavailable RDKit grade; the stopped MDArena unit keeps its original grade.
\end{minipage}
\end{table}

Reported cumulative output totals are 2,721,127 tokens for MDArena and
1,403,186 for AInsteinBench. Usage from the interrupted MDArena request
remains unquantified.
Historical runs used \texttt{deepseek-v4-flash} in August 2026; added attempts
used \texttt{deepseek-flash} in September.

\end{document}